\documentclass{article}
\makeatletter\def\@trackname{}\makeatother
\usepackage[T1]{fontenc}
\usepackage{textcomp}

    \PassOptionsToPackage{numbers, compress}{natbib}

\usepackage[final]{neurips_2025}

\usepackage{soul}
\usepackage[utf8]{inputenc} 
\usepackage[T1]{fontenc}    
\usepackage{hyperref}       
\usepackage{url}            
\usepackage{booktabs}       
\usepackage{amsfonts}       
\usepackage{nicefrac}       
\usepackage{microtype}      
\usepackage{xcolor}         

\usepackage{sidecap}
\usepackage{xcolor}
\usepackage{microtype}
\usepackage{graphicx}
\usepackage{subfigure}
\usepackage{booktabs} 
\usepackage{hyperref}
\usepackage{url}
\usepackage{xurl}
\usepackage{svg}
\usepackage[most]{tcolorbox} 
\usepackage[linesnumbered,ruled,vlined]{algorithm2e}  
\SetKw{Continue}{continue}   

\usepackage{multicol}
\usepackage{wrapfig,lipsum,booktabs}
\usepackage[capbesideposition=outside,capbesidesep=quad]{floatrow}
\hypersetup{
           breaklinks=true,   
           colorlinks=true,   
        }

\usepackage{amsmath}
\usepackage{amssymb}
\usepackage{bm}

\usepackage{mathtools}
\usepackage{amsthm}
\usepackage{float}
\usepackage{enumitem} 
\usepackage{listings}

\tcbset{
  mybox/.style={
    colframe=blue!40!black,     
    breakable,                  
    enhanced,                   
  }
}

\title{Stream: Scaling up Mechanistic Interpretability to Long Context in LLMs via Sparse Attention}

\author{J Rosser\textsuperscript{*}\\
University of Oxford\\
\texttt{jrosser@robots.ox.ac.uk}
 \\
\And
José Luis Redondo García \\
Spotify \\
Spain \\
\And
Gustavo Penha \\
Spotify \\
Netherlands \\
\And
Konstantina Palla \\
Spotify \\
UK \\
\And Hugues Bouchard \\
Spotify \\
Spain \\
}

\begin{document}

\maketitle

\begingroup
\renewcommand\thefootnote{*}
\footnotetext{Work done during an internship at Spotify.}
\endgroup

\begin{abstract}
    As Large Language Models (LLMs) scale to million-token contexts, traditional Mechanistic Interpretability techniques for analyzing attention scale quadratically with context length, demanding terabytes of memory beyond 100,000 tokens. We introduce \textsc{Sparse Tracing}, a novel technique that leverages dynamic sparse attention to efficiently analyze long context attention patterns. We present \textsc{Stream}, a compilable hierarchical pruning algorithm that estimates per-head sparse attention masks in near-linear time $O(T \log T)$ and linear space $O(T)$, enabling one-pass interpretability at scale. \textsc{Stream} performs a binary-search-style refinement to retain only the top-$k$ key blocks per query while preserving the model's next-token behavior. We apply \textsc{Stream} to long chain-of-thought reasoning traces and identify thought anchors while pruning 97-99\% of token interactions. On the RULER benchmark, \textsc{Stream} preserves critical retrieval paths while discarding 90-96\% of interactions and exposes layer-wise routes from the needle to output. Our method offers a practical drop-in tool for analyzing attention patterns and tracing information flow without terabytes of caches. By making long context interpretability feasible on consumer GPUs, \textsc{Sparse Tracing} helps democratize chain-of-thought monitoring. Code is available at \url{https://github.com/spotify-research/stream-mechinterp/}.
\end{abstract}

\section{Introduction}
\label{introduction}


Extending the context length of Large Language Models (LLMs) has become a major research focus with frontier models providing 1M token context length as standard \citep{beltagy2020longformer, zaheer2020big, meta2025llama, comanici2025gemini}. For complex tasks, longer contexts enable longer chain-of-thoughts in reasoning models which can improve performance \citep{jaech2024openai}. In RAG applications, longer context lengths complement retrieval mechanisms, allowing models to consume larger numbers of documents at inference time \citep{li2024retrieval}.

Mechanistic interpretability seeks to reverse engineer neural networks such as LLMs, often by studying smaller toy models such as GPT-2, or decomposing larger models into simpler units that can be analyzed independently \citep{nanda2023attribution}. However, these common techniques such as those introduced by \citet{lindsey2025biology} and \citet{nanda2023attribution} face scaling challenges when applied to long context scenarios. First, meaningful signals become increasingly diffuse as they spread across extended contexts, making it difficult to isolate and detect interpretable patterns.

Second, the attention mechanism's quadratic scaling - $O(T^2)$ in both computational time and memory usage - creates prohibitive resource demands, requiring terabytes of memory to cache all attention patterns for contexts with 100,000 tokens or more.

Scaling efforts in mechanistic interpretability have predominantly focused on accommodating larger model parameters, while the impact of extended context lengths remains significantly underexplored. Many recent studies explicitly defer interpretability analysis of contexts exceeding 100 tokens to future work \citep{paulo2024automatically, lindsey2025biology, templeton2024scaling}, highlighting this critical gap in the field. This paper directly addresses these scalability challenges by introducing an interpretability technique that seamlessly scales to contexts of 100,000 tokens on consumer-grade GPUs, helping democratize long context mechanistic interpretability in realistic settings.

\begin{figure}
    \centering
    \includegraphics[width=\linewidth]{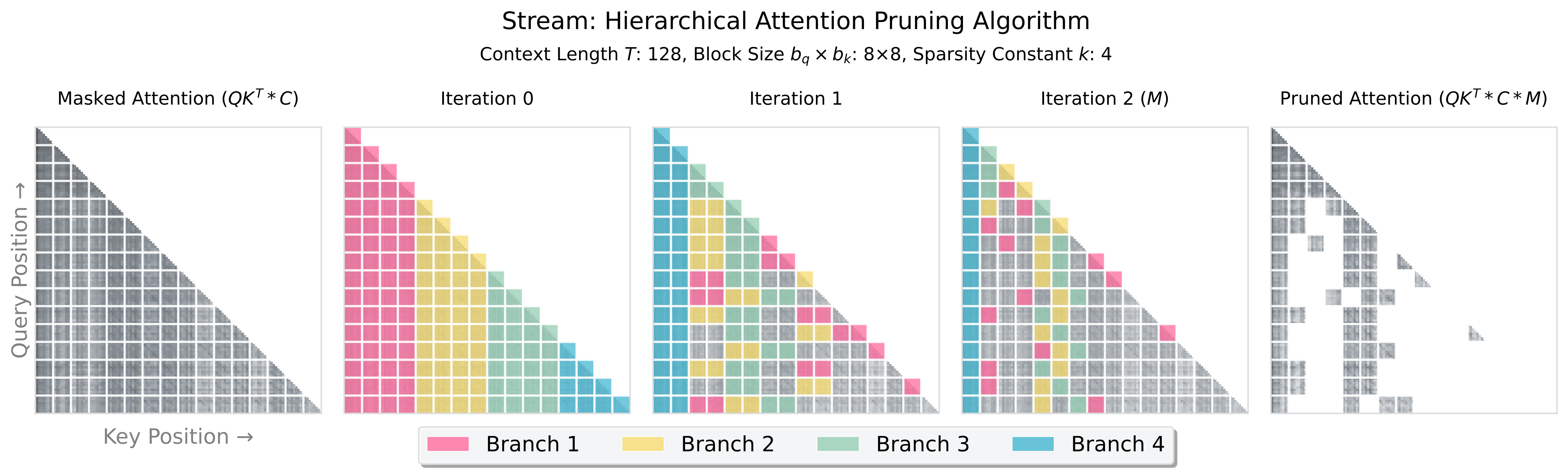}
    \caption{Illustration of the \textsc{Stream} hierarchical attention pruning algorithm. Starting from the full causally masked attention scores ($QK^T * C$), the pattern is recursively divided into branches and refined over successive iterations. At each step, less relevant regions are discarded, converging on a sparse mask $M$ that preserves only the top-$k$ most relevant key-query interactions. The algorithm produces a final pruned attention score pattern ($QK^T*C*M$), achieving near-linear time complexity $O(T log T)$ and linear space complexity $O(T)$.}
    \label{fig:streamgraphic}
\end{figure}

\textbf{Our key contributions are listed as follows:}
\begin{itemize}
    \item \textbf{Scalable long context interpretability.} We introduce \textsc{Sparse Tracing}, the first framework designed to analyze million-token contexts with its instantiation achieving near-linear time $O(T \log T)$ and linear memory $O(T)$, reducing resource costs by up to four orders of magnitude compared to dense methods. 
    
    \item \textbf{Stream: an efficient and flexible algorithm.} We propose \textsc{Stream}, a compilable hierarchical pruning algorithm that dynamically estimates sparse attention masks while allowing fine-grained control over interpretability resolution (e.g. sentence-level vs.\ paragraph-level). 
    
    \item \textbf{Broad validation and impact.} Through chain-of-thought reasoning and RULER benchmarks, we show that \textsc{Stream} prunes 90-99\% of attention links while highlighting critical thought anchors and retrieval paths, enabling long-context mechanistic interpretability on consumer-grade GPUs.
\end{itemize}

\section{Background}
\label{background}


\paragraph{Extending the context length of LLMs.} The standard vanilla self-attention computation on current hardware accelerators is memory-bound, scaling with $O(T^2)$ where $T$ is context length. For example, consider a standard LLM using \texttt{bfloat16} quantization and context length of 100,000 tokens. A single attention pattern would require 20GB of VRAM to realize. Caching all attention patterns across the 34 layers and 8 heads in Gemma 3 4B \citep{team2025gemma} would require 3.84TB. 

FlashAttention \citep{dao2022flashattention} is an exact attention algorithm that is computationally equivalent to vanilla attention but achieves $O(T)$ complexity by chunking the $T\times T$ attention pattern into blocks and computing them sequentially. Other approaches achieve sub-quadratic memory complexity by only computing a subset of each attention pattern, applying a sparse mask in three different ways: (1) the popular static sliding window patterns seen in LLMs designed for long context \citep{beltagy2020longformer, zaheer2020big, team2025gemma}, (2) natively trainable sparse attention (NSA) \citep{yuan2025native} and (3) dynamically generated sparse attention at inference time \citep{lee2024hip, zhang2025spargeattn}.

A lesser challenge LLMs face as context lengths increase is the quadratically scaled memory footprint of the Key-Value (KV) cache during inference. Grouped-Query Attention (GQA) \citep{ainslie2023gqa} allows multiple query heads to share a single key and value head, significantly compressing the cache size with minimal impact on performance. As such, this is often not the bottleneck as context lengths increase. Additionally, long context LLMs are often pretrained on shorter contexts and Rotary Positional Embeddings (RoPE) \citep{su2024roformer} extensions such as YaRN \citep{peng2023yarn},  NTK-Aware interpolation \citep{roziere2023code} and position interpolation \citep{chen2023extending} are used to extend their context length with no impact on time and space complexity \citep{zhong2024understanding}.



\paragraph{Hierarchically Pruned Attention (HiP) Attention.}  \citet{lee2024hip} introduce Hierarchically Pruned (HiP) Attention to address the quadratic scaling challenges introduced by vanilla attention. The transformer’s self-attention layer computes a dense score matrix:
\begin{equation}
S = QK^{\top}, \qquad 
P = \operatorname{softmax}(S), \qquad
O = PV,
\end{equation}
with query, key, and value matrices $Q,K,V \in \mathbb{R}^{T \times d}$ for a context of length $T$.
Because $S$ contains all $T^{2}$ pairwise interactions, both memory use and runtime scale quadratically, quickly becoming the bottleneck for long contexts. The HiP attention algorithm dynamically generates a sparse attention mask at inference time that retains only the top-$k$ keys per query:
\begin{equation}
M = \text{top\_k\_mask}(QK^{\top}), \qquad
\tilde{S} = \operatorname{mask}_{M}(QK^{\top}), \qquad
\tilde{P} = \operatorname{softmax}(\tilde{S}), \qquad
O = \tilde{P}V .
\end{equation}

Here $M \in \{0,1\}^{T \times T}$ is a binary mask with exactly $k$ ones in every row.  
With only $kT$ non-zeros, the softmax and context projection drop to $O(TlogT)$ time and $O(T)$ memory complexity. 
Important hyper-parameters are the sparsity level $k$ and the query and key block sizes $(b_q,b_k)$ that balance locality and throughput. The full HiP attention mask estimation algorithm is provided in Appendix \ref{hip_attention_algorithm} for reference.




\section{Method}
\label{method}


\paragraph{\textsc{Sparse Tracing}; a new technique.} In this paper we introduce \textsc{Sparse Tracing}, a technique comprising a set of mechanistic interpretability algorithms that leverage sparse attention to efficiently analyze attention patterns in long context scenarios. Sparse attention methods reduce computational complexity of the attention computation by computing only the most relevant portions of attention patterns. We hypothesize this may be a useful method of identifying parts of the model related to the model output. Additionally, we observe that the same computational bottlenecks that affect attention computation during inference also hinder interpretability analysis. Additionally, many mechanistic interpretability techniques scale linearly with the number of model components being explored, or require multiple forward and/or backwards passes \citep{nanda2023attribution, kramar2024atp, meng2022locating}. Our technique explores all components at once, and requires only a single forward pass when the desired sparsity constant $k$ is known, and $\left\lfloor log_2k\right\rfloor$ forward passes otherwise, where the wall clock time of a single forward pass with our chosen \textsc{Sparse Tracing} algorithm is less than two forward passes without hooks.

\paragraph{\textsc{Stream}; a new algorithm.} We introduce \textsc{Stream}, our specific \textsc{Sparse Tracing} algorithm based heavily on HiP attention \citep{lee2024hip}.  \textsc{Stream} is visualized in Figure \ref{fig:streamgraphic} and the algorithm is provided in full in Appendix \ref{streamapdx}. HiP attention is a compilable, dynamic sparse attention algorithm that achieves $O(T \log T)$ time complexity and $O(T)$ space complexity and allows the granularity of the attention computation to be tuned. \textsc{Stream} is heavily based on HiP attention and operates via the same hierarchical search process that efficiently identifies the most relevant blocks of keys for each block of queries.

The algorithm begins by dividing the pattern into blocks of size $b_q$ for queries and $b_k$ for keys, then uses a binary search-inspired approach to progressively narrow down to the top-$k$ most relevant key blocks. The pruning process works in iterations, where each iteration refines the search space by half. Initially, \textsc{Stream} divides the entire key pattern into $k$ equally-sized branches. The algorithm then identifies which valid branches contain the highest-scoring attention patterns and discards the less relevant regions. In subsequent iterations, \textsc{Stream} recursively subdivides the remaining promising branches, continuing this hierarchical refinement until it converges on the final set of $k$ key blocks of dimension $b_q \times b_k$ that exhibit the strongest attention relationships with each query block. The behavior of \textsc{Stream} is governed by the 3 key parameters: the query block size $b_q$, the key block size $b_k$, and the sparsity constant $k$.  
Figure \ref{fig:parameters} illustrates how varying these parameters changes the structure of the resulting sparse attention mask.  The granularity of \textsc{Stream} can be adjusted semantically by setting block sizes to approximate linguistic units (e.g., $b_q = b_k = 32$ for sentence-level attention, $b_q = b_k = 128$ for paragraph-level attention). The sparsity level is controlled by setting the sparsity constant $k \in [1, \left\lfloor\frac{T}{b_k}\right\rfloor]$ which controls the level of pruning. Our \textsc{Sparse Tracing} method \textsc{Stream} generates sparse masks for each attention head across all layers. For a given input, we:

\begin{enumerate}
    \item Compute hierarchical sparse attention masks using \textsc{Stream} with specified parameters $(b_q, b_k, k)$.
    \item Apply these masks to the full attention patterns to identify the most relevant attention connections.
    \item Perform binary search over sparsity constant $k$ to find the lowest $k$ that preserves $n_{\text{match}} = 2$ original output tokens, ensuring the sparse pattern captures behaviorally relevant attention.
\end{enumerate}

Our criteria of $n_{\text{match}} = 2$ matching tokens generated in a row is a proxy for perplexity and model performance. We ablate this choice in Appendix \ref{proxyperplex}. \citet{lee2024hip} advise against replacing the first few decoder layers $l_d$ of the transformer with their sparse HiP attention as they found that the first few layers have substantially denser attention patterns and pruning these can degrade performance severely. Through ablation studies they recommend $l_d=3$ which we ablate in Appendix \ref{dense_layers}.

\section{Case Studies}
\label{casestudies}


\subsection{Case Study 1: Thought Anchors}

\citet{bogdan2025thought} introduce black-box and white-box approaches to interpreting long context chain-of-thought reasoning traces by aggregating attention patterns between pairs of sentences rather than between pairs of tokens. They coin the term ``thought anchors'', which are steps in the reasoning trace that have a disproportionately large influence on the remainder of the trace. In this case study, we validate \textsc{Sparse Tracing} as a highly scalable method - in both time and space complexity - of identifying thought anchors.

\paragraph{Experimental Setup.} We follow the experimental setup of \citet{bogdan2025thought}. \textsc{Stream} can be applied to any LLM that implements attention, and for this case study we report results on DeepSeek R1-Distill Qwen-1.5B \citep{guo2025deepseek, yang2024qwen2} a small reasoning in model in the same family as \citet{bogdan2025thought}'s prior work. Matching \citet{bogdan2025thought}, we use a temperature of 0.6 and a top-p value of 0.95 to generate responses to problems in the MATH-500 dataset \citep{hendrycks2021measuring}. In a similar manner to \citet{bogdan2025thought}, we use the results of running the model on 500 problems, 10 times each to identify 10 questions that the model correctly solves 25-75\% of the time. \citet{venhoff2025understanding} identifies a selection of reasoning behaviors exhibited by DeepSeek-R1-Distill models that can be controlled using steering vectors. Based on this framework, \citet{bogdan2025thought} prompt an LLM to label sentences as one of 8 distinct categories. These categories are (1) problem setup, (2) plan generation, (3) fact retrieval, (4) active computation, (5) uncertainty management, (6) result consolidation, (7) self-checking and (8) final answer emission. In our experiments, instead of splitting the reasoning trace into sentences, we split the traces into 32 token chunks to match the query block size ($b_q$) in \textsc{Stream}. We prompt OpenAI GPT-4o (April-May, 2025) \citep{hurst2024gpt} to label each sentence using an almost identical prompt given by \citet{bogdan2025thought} which we provide in Appendix \ref{sentencelabellingprompt}.

\subsubsection{Receiver Heads}

\begin{figure}
    \centering
    \includegraphics[width=\linewidth]{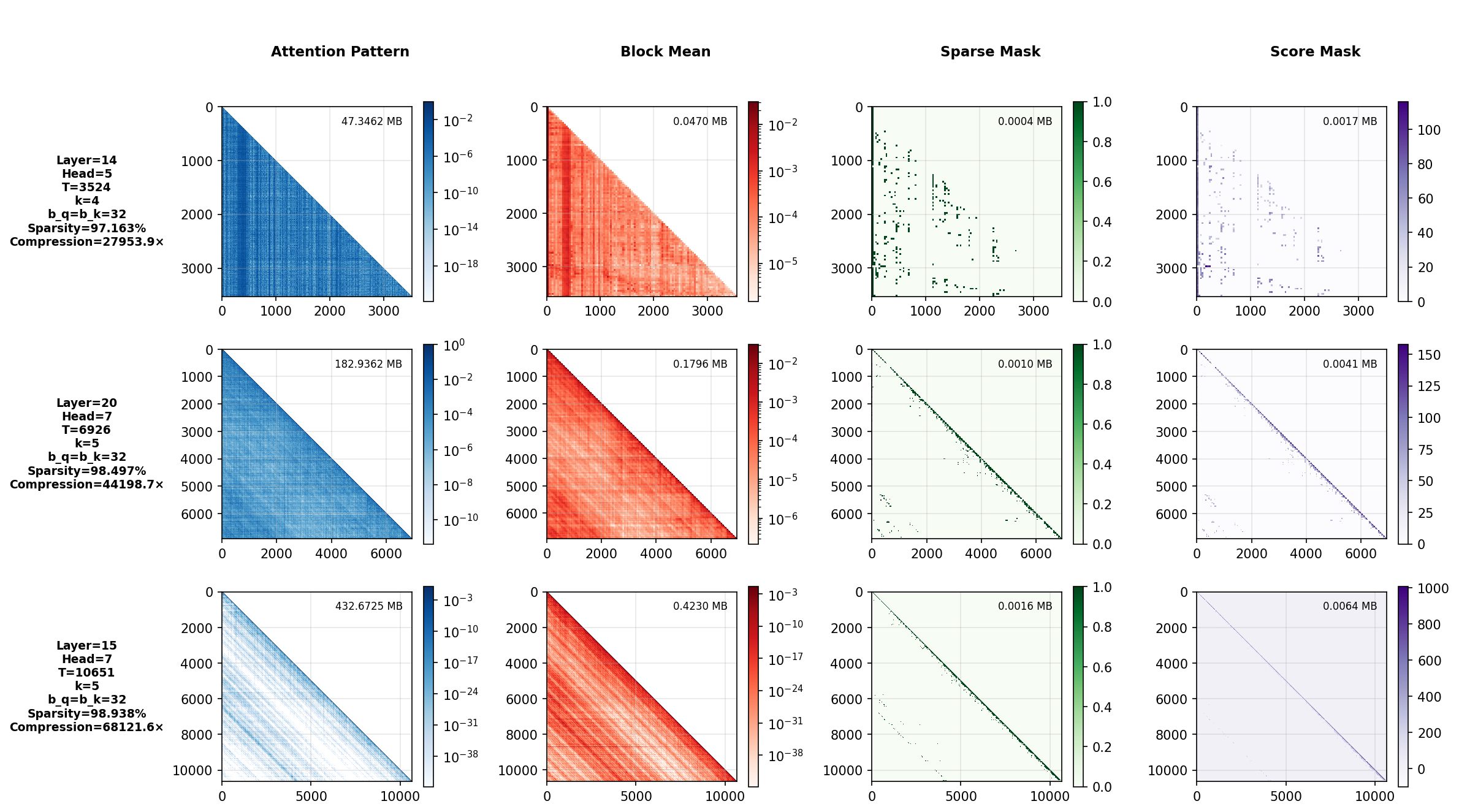}
    \caption{This figure displays attention patterns for three different context lengths (3,000, 6000, and 10000 tokens) using three different approaches: Full Attention Pattern (first column), Mean Averaged \citep{bogdan2025thought} with $b=32$ (second column), and \textsc{Stream} with $b_y=b_k=32$ (third and fourth column). }
    \label{fig:attention_patterns}
\end{figure}

\citet{bogdan2025thought} suggest that ``thought anchors'' attract higher attention from later parts of the text. In \textsc{Sparse Tracing}, we adopt a similar idea: salient query blocks should focus on a small, highly relevant set of key blocks. Formally, let
\begin{equation}
    \mathcal{Q}_i = Q_{\,i : i + b_q - 1}, 
    \qquad
    \mathcal{K}_j = K_{\,j : j + b_k - 1}
\end{equation}
be the $i$th query and $j$th key blocks of sizes $b_q$ and $b_k$. If
\begin{equation}
    \alpha(\mathcal{Q}_i, \mathcal{K}_j) = 
    \max_{u \in \mathcal{Q}_i , v \in \mathcal{K}_j} 
    \langle u, v \rangle
\end{equation}
is the maximum token-level dot product, then
\begin{equation}
    \Pr\!\Big[\alpha(\mathcal{Q}_i, \mathcal{K}_j) 
    \text{ ranks in top-}k\Big]
    \;\propto\; \operatorname{salience}(\mathcal{Q}_i)
\end{equation}
That is, more salient queries appear more often among top-$k$ pairs. To detect such anchors, we look for attention heads with high kurtosis, meaning a few parts of the context receive much stronger attention than the rest. Following \citet{bogdan2025thought}, we call these receiver heads. They concentrate attention on specific blocks instead of spreading it broadly. Instead of aggregating attention to the sentence level, we compute block-level averages over $b_q \times b_k$ spans, yielding the ``block mean'' pattern (Figure \ref{fig:attention_patterns}). We select potential receiver heads by measuring the maximum vertical kurtosis of these block means at multiple context lengths. For each candidate head, we generate sparse masks with \textsc{Stream}, both binary and score-weighted, and visualize them in Figure \ref{fig:attention_patterns}.

\paragraph{Discussion.}  
Across reasoning traces of length 3000, 6000, and 10,000 tokens, \textsc{Sparse Tracing} consistently prunes 97-99\% of attention links while preserving the model’s output for two consecutive tokens. This makes the sparse masks 28,000-68,000$\times$ more memory efficient than storing dense patterns, with complexity scaling linearly in context length (see Appendix \ref{complexity}). Crucially, the peaks in the vertical attention distributions reveal potential thought anchors: specific blocks that disproportionately attract attention from downstream queries. These anchors remain visible despite aggressive pruning, showing that \textsc{Sparse Tracing} highlights the same key structures as the dense patterns but at far lower cost.

\subsubsection{Sentence Category Importance}
\begin{figure}
    \centering
    \includegraphics[width=\linewidth]{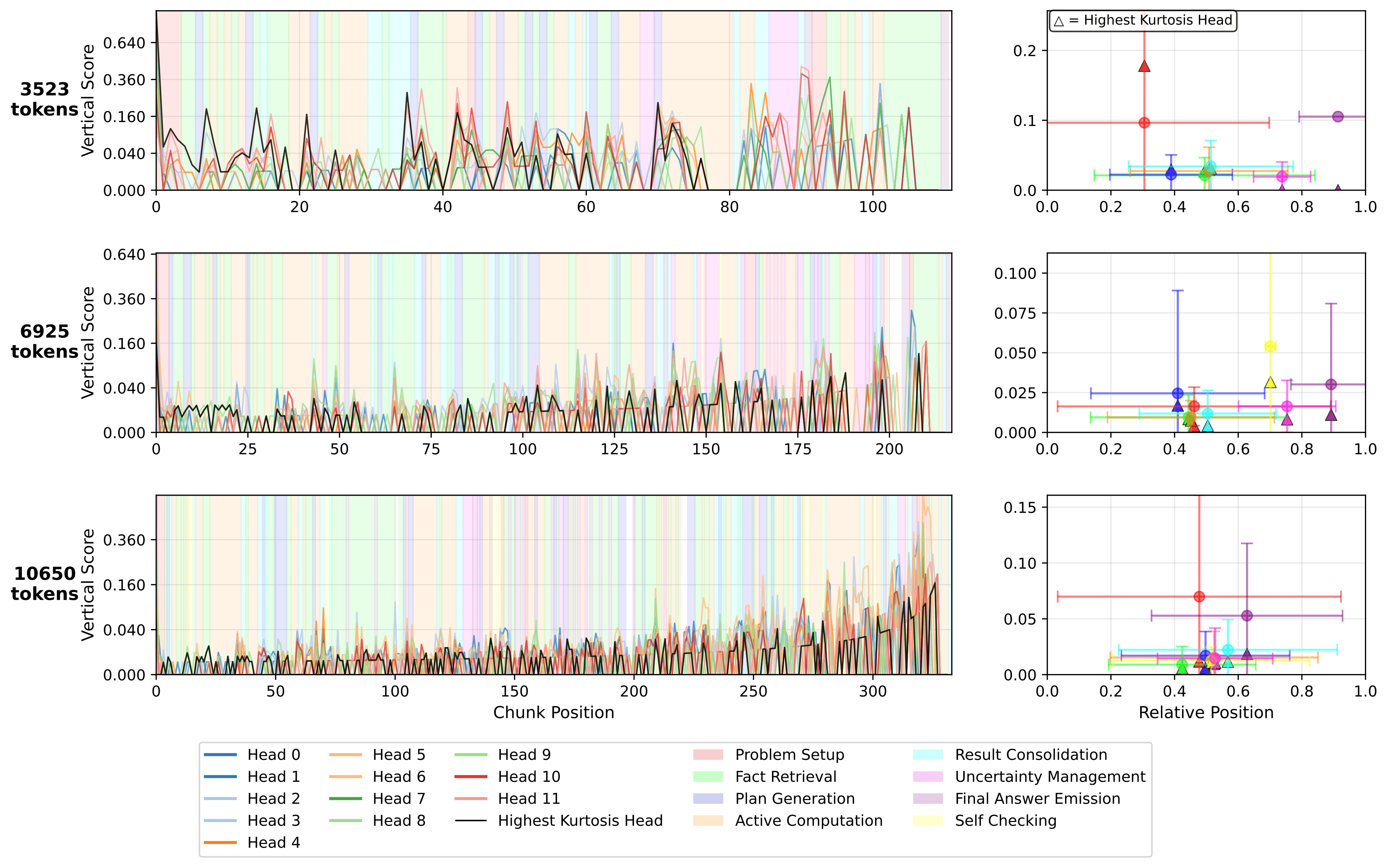}
    \caption{In this figure we plot the vertical attention scores for each 32 token block for three different context lengths (1,000, 10000, and 20,000 tokens).}
    \label{fig:vertical_attention_scores}
\end{figure}

In Figure \ref{fig:vertical_attention_scores}, we compare the sentence category importance based on vertical attention scores, comparing \citet{bogdan2025thought}'s approach and \textsc{Stream}.

\paragraph{Discussion.}  
Figure \ref{fig:vertical_attention_scores} matches the peaks in the vertical scores to identifiable reasoning categories. We observe that problem setup and final answer emission blocks consistently receive the strongest downstream focus, likely due to attention sinks at the beginning and strong diagonal effects near the end of context. In longer traces, receiver heads also peak around plan generation, uncertainty management, and self-checking, while in shorter traces, active computation dominates instead. Overall, we found receiver heads harder to identify as context scales - reporting different heads at different context lengths - and indicating a need for future research to support our results.

\subsection{Case Study 2: Needle in a Haystack}

In this case study, we explore the scalability of \textsc{Stream} by applying it to the RULER needle in a haystack benchmark \citep{hsieh2024ruler}. \textsc{Stream} can be applied to any LLM that implements attention, and we validate the transferability of our method by applying it to a different LLM, namely Gemma 3 1B \citep{team2025gemma}.

\paragraph{Experimental Setup.} The experimental data was prepared using the RULER benchmark's \citep{hsieh2024ruler} synthetic data generation pipeline, which creates structured evaluation samples with controllable context lengths and needle placement positions. For each evaluation instance, we employed a binary search algorithm to determine the optimal sparsity parameter $k$ - the minimum number of key-value blocks required for the sparse attention mechanism to successfully retrieve two consecutive tokens from the needle (see Appendix \ref{proxyperplex} for ablations).

\subsubsection{Sparse Tracing Needles}

\begin{figure}
    \centering
    \includegraphics[width=\linewidth]{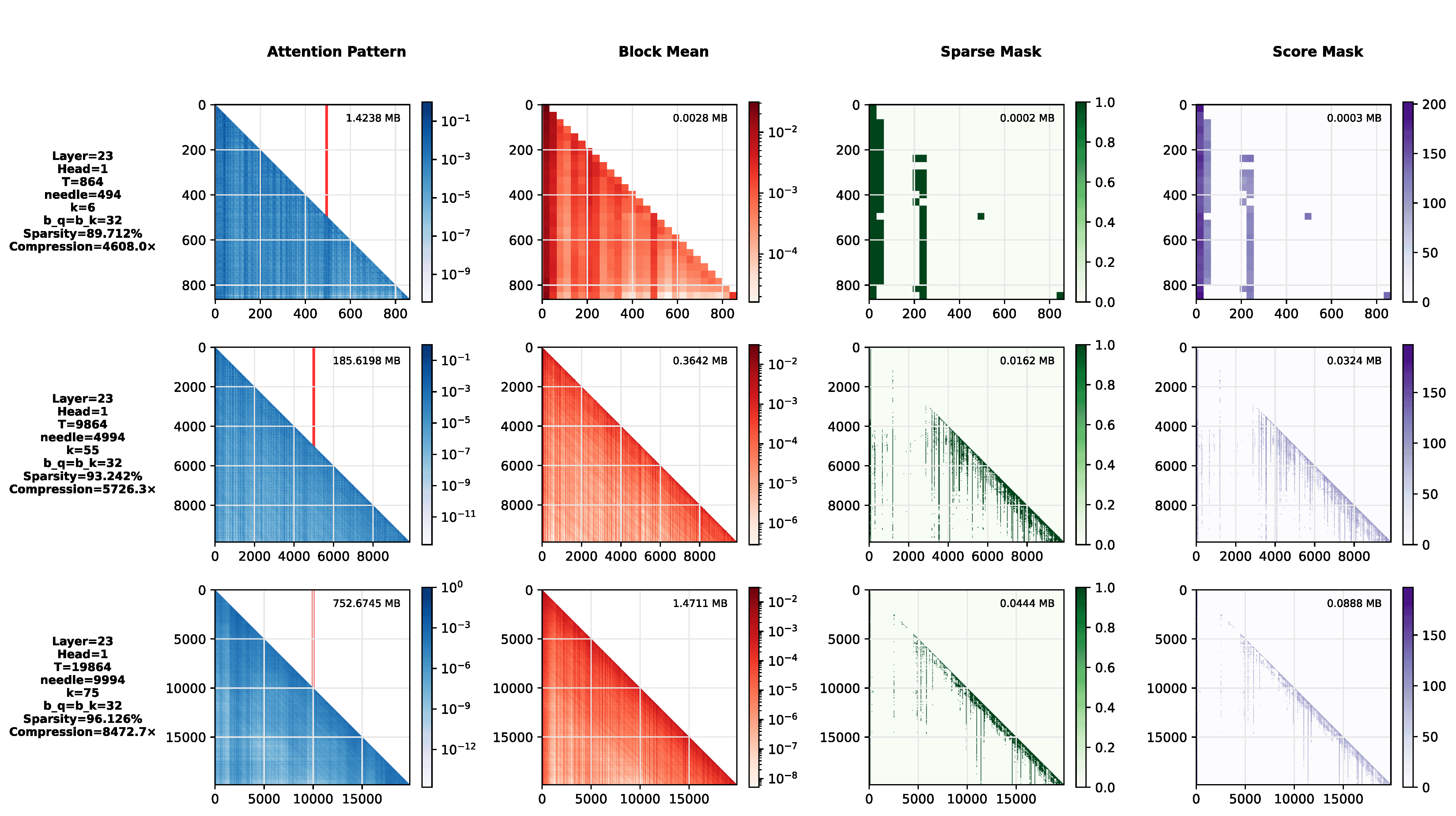}
    \caption{Visualization of sparse attention masks generated by \textsc{Stream} on the RULER needle-in-a-haystack benchmark \citep{hsieh2024ruler} at three context lengths (1k, 10k, and 20k tokens). The needle is marked using a red vertical line in the Attention Pattern plots.}
    \label{fig:niah}
\end{figure}

In Figure \ref{fig:niah} we visualize the needle at 3 different context lengths (1000, 10000 and 20000 tokens). We extract the full attention patterns from a Gemma 3 1B model at the layer and attention head with highest block mean kurtosis.

\paragraph{Discussion.} The resulting sparse masks are strikingly structured: despite discarding over 90-96\% of attention links, the critical path to the needle remains intact and is clearly visible in the pruned attention maps. This demonstrates \textsc{Stream}'s ability to aggressively reduce computational complexity while still preserving the essential signal required for successful retrieval, even as context length scales by an order of magnitude.

\subsubsection{Exploring the effect of Needle Depth}

\begin{figure}
    \centering
    \includegraphics[width=\linewidth]{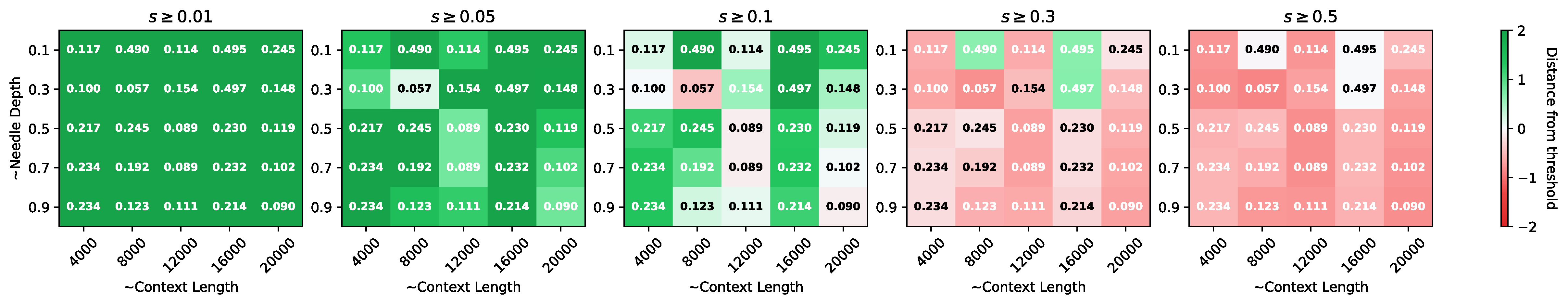}
    \caption{Figure showing needle retrieval success as a function of context length (x-axis), needle depth (y-axis) and effective sparsity $s$ (subplots-axis). The parameter $s\in(0,1)$ controls the effective pruning of the attention pattern via $k$ with larger $s$ corresponding to higher pruning.}
    \label{fig:niah_depth}
\end{figure}

In needle-in-a-haystack tasks, models often show a U-shaped retrieval curve \citep{liu2023lost}: they are relatively good at retrieving needles near the start of the context (due to over-squashing) and near the end (as those positions are directly learned), but struggle in the middle. By introducing the effective sparsity $s$ such that:
\begin{equation}
    k = \left\lfloor 1+ \bigg(\frac{T}{b_q} -1\bigg)\cdot s \right\rfloor \quad \text{such that} \quad 1\leq k \leq \left\lfloor \frac{T}{b_q}\right\rfloor \quad \text{and} \quad 0<s<1
\end{equation}
The larger the $s$, the higher the pruning of the attention pattern. Figure \ref{fig:niah_depth} visualizes successful needle retrieval as $s$, context length and needle depth vary. We see that the retrieval of needles earlier in the context is more robust as context length scales.

\paragraph{Discussion.} Instead of the characteristic U-shape, we find that \textsc{Stream}'s attention pruning weakens needle retrieval towards the end of longer contexts with effective sparsity $s \geq 0.1$, leading to failures above 8000 tokens. The lower triangular nature of causal attention patterns results in a larger number of valid candidate key blocks per query towards the bottom of the pattern. This leads to a higher level of pruning at constant $k$, which would be resolved by implementing variable $k$ in future research.

\subsubsection{Exploring Information Flow}
\begin{figure}
    \centering
    \includegraphics[width=\linewidth]{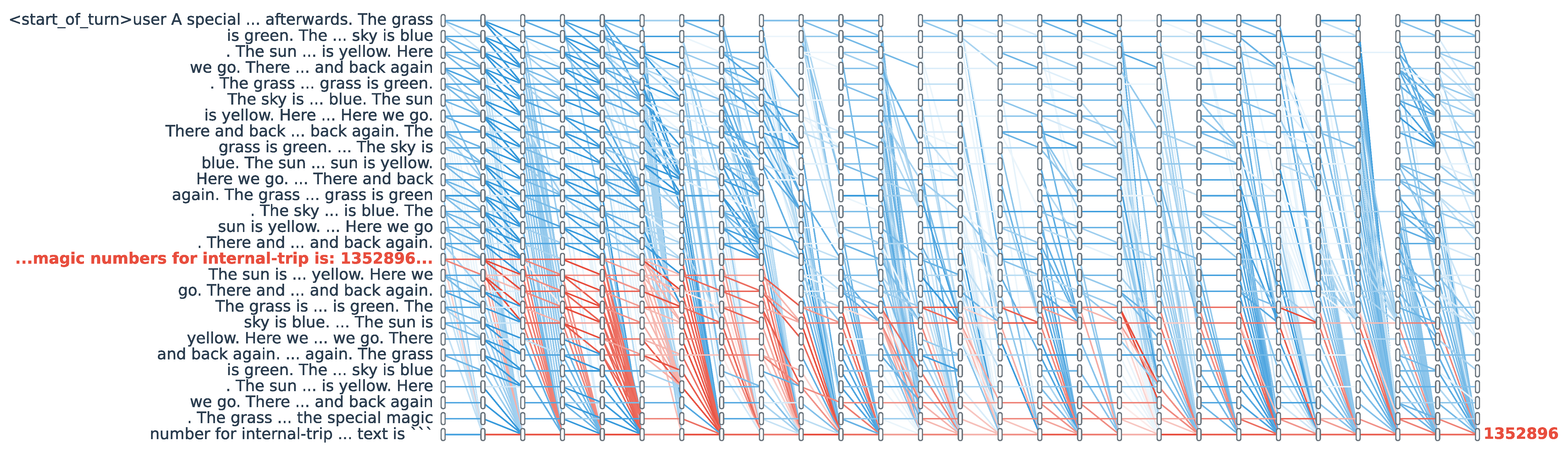}
    \caption{Attention flow paths distinguishing successful and unsuccessful retrieval in the needle-in-a-haystack task ($T=1000$, $n_d \approx 0.5$). Each node represents a hidden state block of size $b_q=b_k=32$. Red edges trace paths carrying needle information to the output, while blue edges denote other attention flows. Paths are weighted by block scores of $k_{\text{success}}$, and obtained by subtracting the sparse masks of $k_{\text{fail}}=3$ from $k_{\text{success}}=6$.}
    \label{fig:flowsubtract}
\end{figure}

Existing methods for tracing information flow in transformers \citep{ferrando2024information, nanda2023attribution, kramar2024atp, meng2022locating} scale quadratically with context length. To our knowledge, \textsc{Stream} is the first such technique with linear scaling, making it feasible for tasks like indirect object identification and needle-in-a-haystack. For the latter, we report sparse patterns at $T=1000$, needle depth $n_d \approx 0.5$, $b_q=b_k=32$, with $k_{\text{success}}=6$ (successful retrieval) and $k_{\text{fail}}=3$ (unsuccessful retrieval). Subtracting the $k_{\text{fail}}$ masks from $k_{\text{success}}$ across layers and heads isolates the connections that enable retrieval. Figure \ref{fig:flowsubtract} visualizes these paths, weighted by block scores: red edges trace needle-to-output flows, while blue edges denote other attention routes. We restrict analysis to attention paths, leaving residual and MLP components for future \textsc{Sparse Tracing} extensions. Full patterns for $k_{\text{success}}$ and $k_{\text{fail}}$ are shown in Appendix \ref{infapp}.

\paragraph{Discussion.} We see that the needle information appears to travel along two paths to the output, through hidden state blocks 18 and 19, and through the final few blocks. We also see attention sink behavior - high attention to earlier tokens in the context - when needle information passes to the output block, which may be preventing over-mixing, a form of rank collapse \citep{barbero2025llms}. This would aid in preserving the information from the needle through the layers of the network. This experiment consolidates the use of \textsc{Stream} as an empirical tool to explore theoretical claims.

\section{Related Work}
\label{relatedwork}


\paragraph{Chain-of-Thought White-box Monitoring.} Reasoning models produce a long chain of thought before responding to the user, with the aim of increasing their performance on complex tasks \citep{jaech2024openai}. A growing branch of Mechanistic Interpretability research seeks to understand model internals during chain-of-thought processes \citep{bogdan2025thought, cabannes2024iteration, brinkmann2024mechanistic, dutta2024think, venhoff2025understanding}. \citet{brinkmann2024mechanistic} demonstrate that transformers can implement interpretable backward chaining algorithms for pathfinding, while \citet{dutta2024think} extend this understanding to production-scale models, revealing multiple parallel pathways for answer generation. \citet{venhoff2025understanding} show these reasoning behaviors can be controlled through steering vectors. \citet{cabannes2024iteration} find identifiable computational structures that enable iterative algorithms, and \citet{bogdan2025thought} take black-box and white-box approaches to identify "thought anchors" - the most critical parts of the reasoning trace. In our work, we use our \textsc{Sparse Tracing} methodology to scalably identify "thought anchors" in longer contexts.

\paragraph{Why LLMs struggle with Long Context.} Other work examines how LLMs process extended contexts and the mechanisms underlying their successes and failures \citep{wu2024retrieval, liu2023lost, barbero2024round, barbero2024transformers, liu2025survey}. \citet{wu2024retrieval} identify "retrieval heads" - a sparse set of attention heads (3-6\%) specifically responsible for copying information from arbitrary locations in long contexts, providing a mechanistic explanation for factual retrieval capabilities. This work builds on \citet{liu2023lost}'s empirical finding that models perform poorly when relevant information appears in the middle of long contexts. \citet{barbero2024transformers} provide theoretical grounding for these limitations, proving that decoder-only transformers suffer from "information over-squashing" where distinct input contexts become indistinguishable in final token representations. Their companion work \citep{barbero2024round} reveals how rotary positional embeddings can mitigate some issues by using different frequency components for positional versus semantic attention. Our work introduces \textsc{Stream}, which builds highly scalable computation graphs for information propagation research into Geometric Deep Learning metrics such as over-smoothing, over-squashing and under-reaching.

\paragraph{Scaling Mechanistic Interpretability.} Tooling has been built for conducting Mechanistic Interpretability research as model size, complexity and context lengths scale \citep{nanda2023attribution, kramar2024atp, ferrando2024information, zheng2024attention}. \citet{nanda2023attribution} introduces attribution patching, enabling activation patching experiments on large models through gradient-based approximations. \citet{kramar2024atp} address failure modes in this approach, proposing AtP*. \citet{ferrando2024information} develop automated methods for discovering information flow routes without requiring human-designed counterfactuals, while \citet{zheng2024attention} provide systematic frameworks for understanding attention head functions. \citet{paulo2024automatically} address the scalability challenge of interpreting sparse autoencoder (SAE) features by developing an automated pipeline that can generate and evaluate natural language explanations for millions of features. \citet{templeton2024scaling} successfully scale SAEs to Claude 3 Sonnet extracting millions of interpretable features. We introduce the first Mechanistic Interpretability tool designed to run on consumer hardware that scales to long context.

\section{Limitations}
\label{limitations}


(1) \textsc{Stream} focuses on the attention mechanism and makes no causal claims. Residual connections and MLP layers also play crucial roles in computation, and \textsc{Stream} does not currently account for these. (2) The method's success is defined by a proxy metric, specifically maintaining two consecutive correct tokens in the output. This is a practical heuristic but may not fully capture the nuance of model performance or the complete causal path. (3) While \textsc{Stream} achieves impressive computational complexity which holds as contexts scale, our current implementation has not been optimized for specific hardware and therefore runs slower than desired. (4) The pruning algorithm is more aggressive towards the bottom of the attention pattern, which could be resolved by implementing a variable sparsity constant $k$. (5) Future work introducing a novel discovery made by \textsc{Stream} would truly validate our method's applicability.

\section{Conclusion}
\label{conclusion}


Our work introduces \textsc{Sparse Tracing} and its instantiation, \textsc{Stream}, as scalable tools for mechanistic interpretability in long context LLMs. These methods address a key limitation of existing interpretability techniques: the quadratic complexity of attention, which makes direct analysis of million-token contexts practically infeasible. By leveraging hierarchical pruning, \textsc{Stream} achieves near-linear $O(T\log T)$ time and linear $O(T)$ space complexity while preserving salient attention pathways, enabling experiments previously restricted to short contexts. We apply \textsc{Stream} to both reasoning (DeepSeek R1-Distill Qwen-1.5B) and non-reasoning models (Gemma 3 1B) and through 5 experiments and 3 ablations, we validate the widespread applicability of our method. In chain-of-thought reasoning at 10,000 token context, \textsc{Sparse Tracing} prunes 99\% of the attention pattern and identifies influential ``thought anchors'' requiring 60,000$\times$ less memory than caching the full patterns. In needle-in-a-haystack tasks, \textsc{Stream} highlights the sparse structures required for needle retrieval, pruning the pattern by 96\% at 20,000 token context. 

\paragraph{Future Work. } (1) Future work should incorporate MLP layers and residual stream analysis into the \textsc{Sparse Tracing} framework providing a more complete picture of information flow. (2) Establishing theoretical guarantees about what types of circuits and information pathways are preserved under different sparsity levels would strengthen confidence in sparse interpretability results. (3) \textsc{Stream} opens new possibilities for studying long context phenomena like information over-squashing, over-smoothing, and under-reaching. (4) Further work could also explore the effect of aligning block sizes with semantic units such as sentences or paragraphs and extend \textsc{Stream} to further case studies.

\begin{ack}
J Rosser is supported by the EPSRC Centre for Doctoral Training in Autonomous and Intelligent Machines and Systems EP/Y035070/1. This work was carried out during an internship at Spotify, who also provided the computational resources. We are especially grateful to Federico Barbero, whose insight and early guidance were instrumental in shaping the direction of this research, and who provided invaluable technical input throughout the project. We also thank Mounia Lalmas-Roelleke, Paul Bennett, Ali Vardasbi, Enrico Palumbo, and Jiazheng Li for generously providing thoughtful feedback and discussions.
\end{ack}

\bibliographystyle{plainnat}
\bibliography{neurips_2025.bib}

\appendix
\pagebreak

\section{Ablations}
\label{ablations}

\subsection{Ablation Study of Dense Layers}
\label{dense_layers}

\begin{figure}[H]
    \centering
    \includegraphics[width=\linewidth]{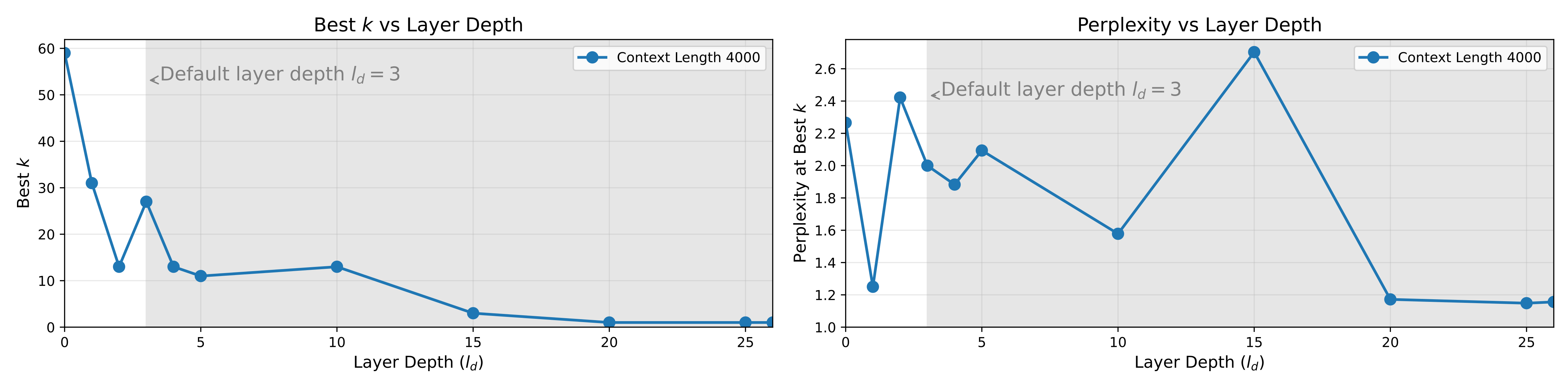}
    \caption{Ablation of dense layers. Left: best sparsity constant $k$ as a function of layer depth. Right: perplexity at the corresponding best $k$. Results support the choice of leaving the first $l_d=3$ layers dense, where attention patterns are less sparse and pruning leads to degraded performance.}
    \label{fig:dense_ablation}
\end{figure}

To validate $l_d=3$ , we perform a similar ablation study as in Appendix D.5 in \citet{lee2024hip}'s study and we report results in Figure \ref{fig:dense_ablation}.

\subsection{Ablation Study of Number of Consecutive Tokens Matching as a Proxy for Perplexity}
\label{proxyperplex}

\begin{figure}[H]
    \centering
    \includegraphics[width=\linewidth]{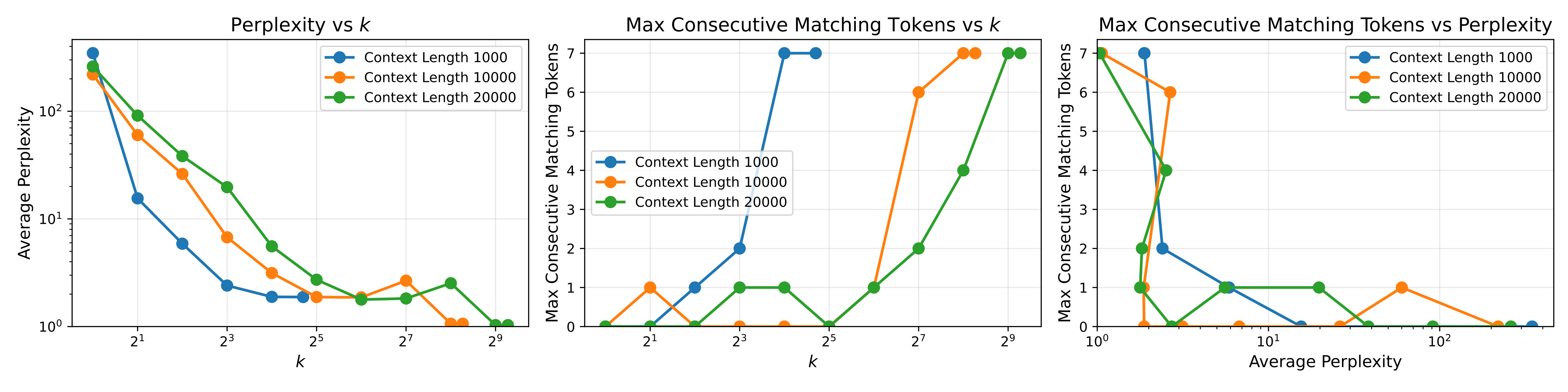}
    \caption{Ablation of sparsity constant $k$ search. Left: average perplexity decreases with increasing $k$. Middle: number of consecutive matching tokens improves with larger $k$, with $n_{\text{match}}=2$ serving as our performance threshold. Once this threshold is met, the model typically produces the full correct output. Right: trade-off between perplexity and matching tokens highlights the balance between pruning aggressiveness and output fidelity.}
    \label{fig:consecutive}
\end{figure}

To maximize pruning efficiency, we aim to select the smallest sparsity constant 
$k$ that still preserves model performance. Directly measuring degradation via average perplexity is impractical, since it is unclear at what threshold perplexity becomes unacceptable. Instead, we adopt a behavioral proxy: running the model with sparse masks for varying $k$ and checking when it can no longer reproduce the same output. We define performance preservation as generating at least $n_{\text{match}}=2$ consecutive matching tokens, and report ablations for this criterion in Figure \ref{fig:consecutive}. Notably, in the middle and right plots, we observe that once $k$ or average perplexity is large enough to achieve $n_{\text{match}}=2$, at any $k$ higher the model typically recovers the entire output sequence, suggesting that this threshold is a reliable indicator of stable performance.  

\subsection{Ablation Study of Time and Space Complexity}
\label{complexity}

\begin{figure}[H]
    \centering
    \includegraphics[width=\linewidth]{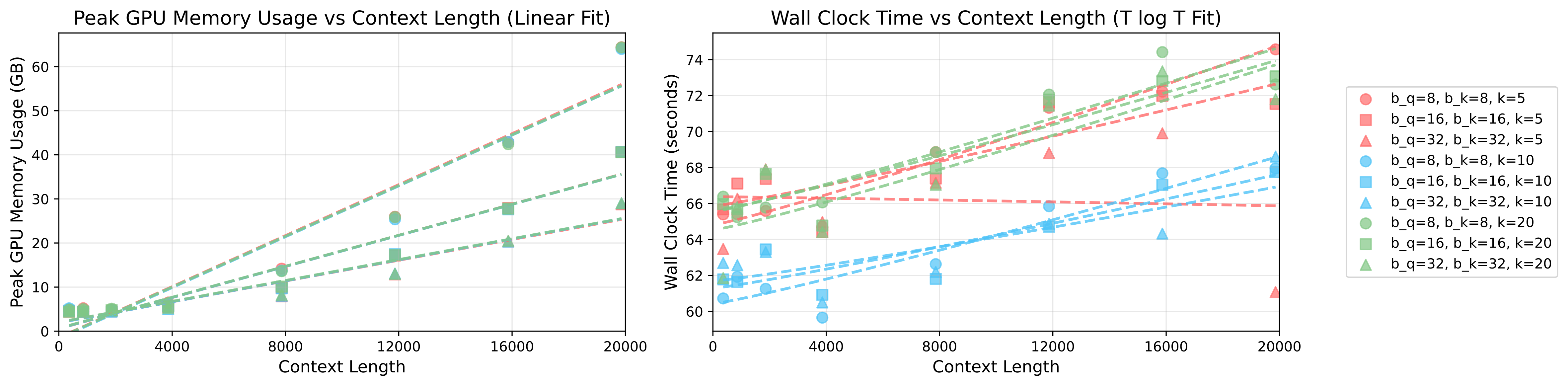}
    \caption{GPU memory usage (linear fits, mean $R^2$ = 0.9308) and wall clock time (T log T fits, mean $R^2$ = 0.7282) versus context length for varying block sizes ($b_q$, $b_k$) and sparsity levels $k$.}
    \label{fig:memory}
\end{figure}

In Figure \ref{fig:memory} we report the GPU Memory Usage and Wall Clock Time as they vary with sparsity level $k$, context length and block size ($b_q$, $b_k$).

\section{HiP Attention Algorithm}
\label{hip_attention_algorithm}

In Algorithm \ref{maskestimation} we provide the full  Hierarchical Sparse Attention Mask Estimation algorithm as presented by \citet{lee2024hip}.

\begin{algorithm}[H]
\caption{Hierarchical Sparse Attention Mask Estimation}\label{maskestimation}
\KwIn{Queries $\bm{Q} \in \mathbb{R}^{T\times d}$, Keys $\bm{K} \in \mathbb{R}^{T\times d}$, Sparsity constant $k$, Query block size $b_q$, Key block size $b_k$, Top-$r$ approximation constant $r$}
\KwOut{Estimated attention mask $\widehat{\bm{M}} \in \{0, 1\}^{T\times T}$ which is represented by an array of indices $\mathcal{I} \in [1:T]^{T/b_q \times k/b_k}$}

$\bm{\mathsf{Q}}, \bm{\mathsf{K}} = \mathrm{reshape}_{T/b_q\times b_q\times d}[\bm{Q}], \mathrm{reshape}_{T/b_k\times b_k\times d}[\bm{K}]$\;
$n_{it} = \lceil \log(T/b_k)\rceil$\; \tcp{Number of iterations}

\For{each query block index $q = 1~..~T/b_q$}{
    $\left(f_{qj}^{(1)}, l_{qj}^{(1)}\right) = \left(\lfloor j\cdot \frac{T}{k} \rfloor, \lfloor (j + 1)\cdot \frac{T}{k} \rfloor - 1\right)$ for $j = 1~..~k$\; \tcp{Set $k$ nodes' initial start and end indices}
    
    \For{each iteration $i = 1~..~n_{it}$}{
        \For{each node index $j = 1~..~k$}{
            $m^{(i)}_{qj} = \lfloor (f^{(i)}_{qj} + l^{(i)}_{qj})/2 \rfloor$\;
            $\left( \mathcal{B}^{(i)}_{q,2j-1}, \mathcal{B}^{(i)}_{q,2j} \right) = \left( (f^{(i)}_{qj} : m^{(i)}_{qj} - 1), (m^{(i)}_{qj} : l^{(i)}_{qj}) \right)$\;
        }
        
        \For{each branch index $h = 1~..~2k$}{
            Pick a first index $r^{(i)}_{qh}$ from the range $\mathcal{B}^{(i)}_{qh}$\;
            Compute score $s^{(i)}_{qh} = \max_{m,n}\left(\bm{\mathsf{Q}}_{q,m,:} ^\top \bm{\mathsf{K}}_{r^{(i)}_{qh},n,:}\right)$\;
        }
        
        Pick top-$k$ indices $\{t_{1}, \dots, t_{k}\}$ of the sequence $s^{(i)}_{q,1}, \dots, s^{(i)}_{q,2k}$\;
        Update nodes $(f^{(i+1)}_{qj}: l^{(i+1)}_{qj}) := \mathcal{B}^{(i)}_{t_j}$ for $j = 1~..~k$\;
    }
    
    Set mask indices $\mathcal{I}_{qj} = f^{(n_{it})}_{qj}$ for $j = 1~..~k$\;
}
\end{algorithm}

\section{\textsc{Stream} Algorithm}
\label{streamapdx}

\begin{figure}[H]
    \centering
    \includegraphics[width=\linewidth]{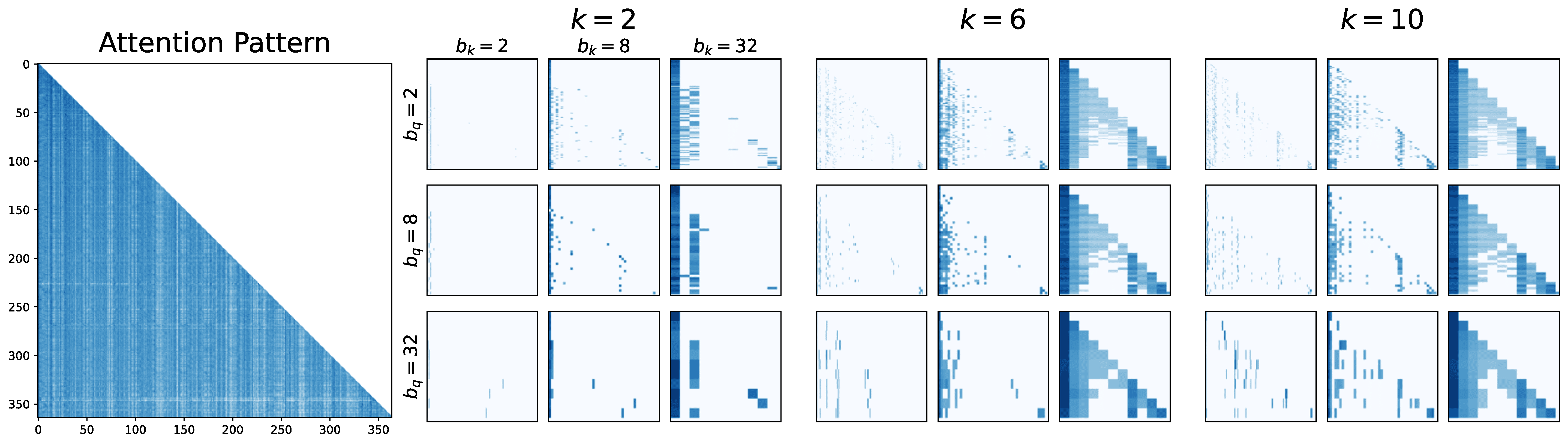}
    \caption{Effect of block and sparsity parameters on sparse attention masks.  
    Increasing $b_q$ and $b_k$ coarsens the mask by aggregating larger query and key spans, while higher values of $k$ preserve more key blocks, yielding denser masks.}
    \label{fig:parameters}
\end{figure}

We provide the full \textsc{Stream} algorithm in Algorithm \ref{alg:stream}. In Figure \ref{fig:parameters} we visualize the effect of $b_q$, $b_k$ and $k$ on the final sparse attention mask.

\begin{algorithm}[H]
\caption{\textsc{Stream}}\label{alg:stream}
\KwIn{Queries $\bm{Q}\!\in\!\mathbb{R}^{T\times d}$, Keys $\bm{K}\!\in\!\mathbb{R}^{T\times d}$, Attention mask $\bm{C}\!\in\!\{0,1\}^{T\times T}$, Sparsity $k$, Query block $b_q$, Key block $b_k$, Top-$r$ constant $r$}
\KwOut{$\widehat{\bm{M}}\!\in\!\{0,1\}^{T\times T}$ via indices $\mathcal{I}\!\in\![1:T]^{T/b_q \times k/b_k}$}

$\ell \gets \text{lcm}(b_q, b_k)$\;
$T_p \gets \lceil T / \ell \rceil \cdot \ell$\;
$\text{extra} \gets T_p - T$\;
$T \gets T_p$\;
$Q\gets \text{pad}(Q, ((0, \text{extra}), (0, 0))), \quad K \gets  \text{pad}(K, ((0, \text{extra}), (0, 0)))$\;

$C \gets \text{pad}(C, ((0, \text{extra}), (0, \text{extra})))$

$\bm{\mathsf{Q}},\bm{\mathsf{K}}=\mathrm{reshape}_{T/b_q\times b_q\times d}[\bm{Q}],~\mathrm{reshape}_{T/b_k\times b_k\times d}[\bm{K}]$\;
$\bm{\mathsf{C}}=\mathrm{reshape}_{T/b_q\times b_q\times T/b_k\times b_k}[\bm{C}]$\;
$\bm{C}^{\mathrm{blk}}\in\{0,1\}^{T/b_q\times T/b_k},~\bm{C}^{\mathrm{blk}}[q,r]=\max_{m\le b_q,n\le b_k}\bm{\mathsf{C}}[q,m,r,n]$\;
$n_{it}=\lceil\log(T/b_k)\rceil$\;

\For{$q=1~..~T/b_q$}{
  $\big(f_{qj}^{(1)},l_{qj}^{(1)}\big)=\big(\lfloor j\frac{T}{k}\rfloor,\ \lfloor (j+1)\frac{T}{k}\rfloor-1\big)$ for $j=1~..~k$\;
  \For{$i=1~..~n_{it}$}{
    \For{$j=1~..~k$}{
      $m^{(i)}_{qj}=\lfloor(f^{(i)}_{qj}+l^{(i)}_{qj})/2\rfloor$\;
      $\big(\mathcal{B}^{(i)}_{q,2j-1},\mathcal{B}^{(i)}_{q,2j}\big)=\big((f^{(i)}_{qj}:m^{(i)}_{qj}-1),(m^{(i)}_{qj}:l^{(i)}_{qj})\big)$\;
    }
    \For{$h=1~..~2k$}{
      $\omega^{(i)}_{qh}=\max\limits_{r\in\mathcal{B}^{(i)}_{qh}}\bm{C}^{\mathrm{blk}}[q,r]$\;
      \uIf{$\omega^{(i)}_{qh}=0$}{set $s^{(i)}_{qh}=-\infty$ and \textbf{continue}}
      Pick first $r^{(i)}_{qh}\in\mathcal{B}^{(i)}_{qh}$ with $\bm{C}^{\mathrm{blk}}[q,r^{(i)}_{qh}]=1$\;
      Compute $s^{(i)}_{qh}=\max\limits_{\substack{m\le b_q,\,n\le b_k\\ \bm{\mathsf{C}}[q,m,r^{(i)}_{qh},n]=1}}\!\!\bm{\mathsf{Q}}_{q,m,:}^{\top}\bm{\mathsf{K}}_{r^{(i)}_{qh},n,:}$\;
    }
    Pick top-$k$ indices $\{t_1,\dots,t_k\}$ of $s^{(i)}_{q,1},\dots,s^{(i)}_{q,2k}$\;
    Update $(f^{(i+1)}_{qj}:l^{(i+1)}_{qj}) := \mathcal{B}^{(i)}_{t_j}$ for $j=1~..~k$\;
  }
  Set $\mathcal{I}_{qj}=$ first $r\in(f^{(n_{it})}_{qj}:l^{(n_{it})}_{qj})$ with $\bm{C}^{\mathrm{blk}}[q,r]=1$; if none, leave $\mathcal{I}_{qj}$ empty\;
}
\end{algorithm}

\section{Thought Anchors: Additional Results}

\subsection{Sentence Labelling Prompt}
\label{sentencelabellingprompt}

Here we repeat the prompt provided by \citet{bogdan2025thought} to label sentences, with minor modifications.

{
\tt You are an expert in interpreting how language models solve math problems using multi-step reasoning. 
Your task is to analyze a Chain-of-Thought (CoT) reasoning trace, broken into discrete text chunks, and 
label each chunk with the single best **function\_tag** that describes what the chunk is doing.

---

\#\#\# Function Tags (you may only assign one per chunk):

1. `problem\_setup`: 
    Parsing or rephrasing the problem (initial reading or comprehension).
    
2. `plan\_generation`: 
    Stating or deciding on a plan of action (often meta-reasoning).
    
3. `fact\_retrieval`: 
    Recalling facts, formulas, problem details (without immediate computation).
    
4. `active\_computation`: 
    Performing algebra, calculations, manipulations toward the answer.
    
5. `result\_consolidation`: 
    Aggregating intermediate results, summarizing, or preparing final answer.
    
6. `uncertainty\_management`: 
    Expressing confusion, re-evaluating, proposing alternative plans (includes backtracking).
    
7. `final\_answer\_emission`: 
    Explicit statement of the final boxed answer or earlier chunks that contain the final answer.
    
8. `self\_checking`: 
    Verifying previous steps, Pythagorean checking, re-confirmations.

9. `unknown`: 
    Use only if the chunk does not fit any of the above tags or is purely stylistic or semantic.
    
---

\#\#\# Output Format:

Return a single dictionary with one entry per chunk, where each entry has:
- the chunk index (as the key, converted to a string),
- a string with the tag

Here's the expected format:

```language=json\\
\{\\
    "0": "plan\_generation"\\
    "1": "fact\_retrieval"\\
    "2": "active\_computation"\\
    "3": "self\_checking"\\
    "4": "final\_answer\_emission"\\
    ...\\
\}\\
```\\
Here is the math problem:\\

[PROBLEM]\\
\{problem\_text\}

Here is the full Chain of Thought, broken into chunks:\\

[CHUNKS]\\
\{full\_chunked\_text\}

Now label each chunk with function tags only.
}

\subsection{Full Attention Comparison}

We extend Figure \ref{fig:attention_patterns} and include the vertical attention scores for block mean, sparse mask and score mask for the receiver head.

\begin{figure}[H]
    \centering
    \includegraphics[width=0.9\linewidth]{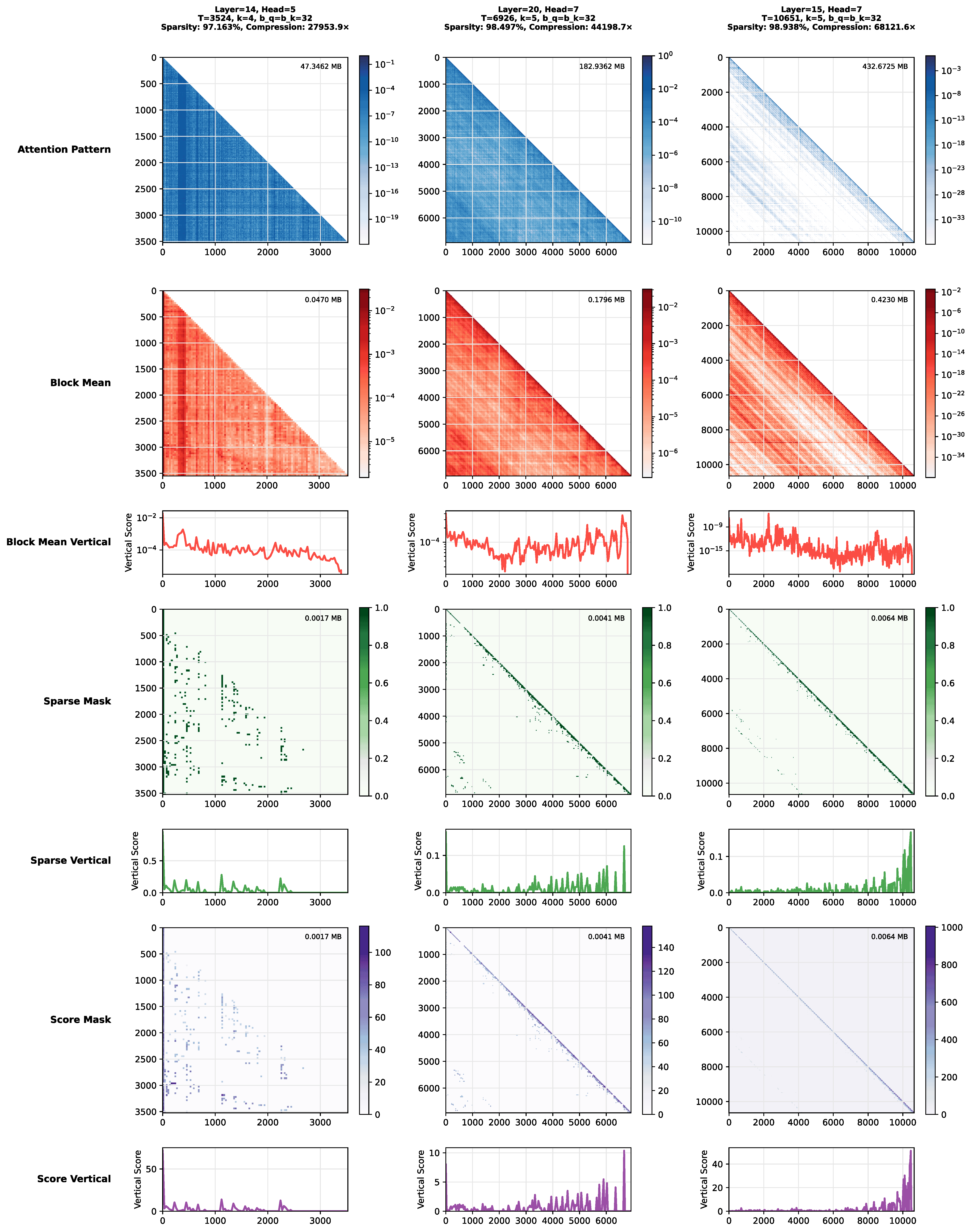}
    \caption{Extending Figure \ref{fig:attention_patterns} to include the vertical attentions score plots.}
    \label{fig:fullatt}
\end{figure}

\subsection{Sentence Category Distribution}
\begin{figure}[H]
    \centering
    \includegraphics[width=0.7\linewidth]{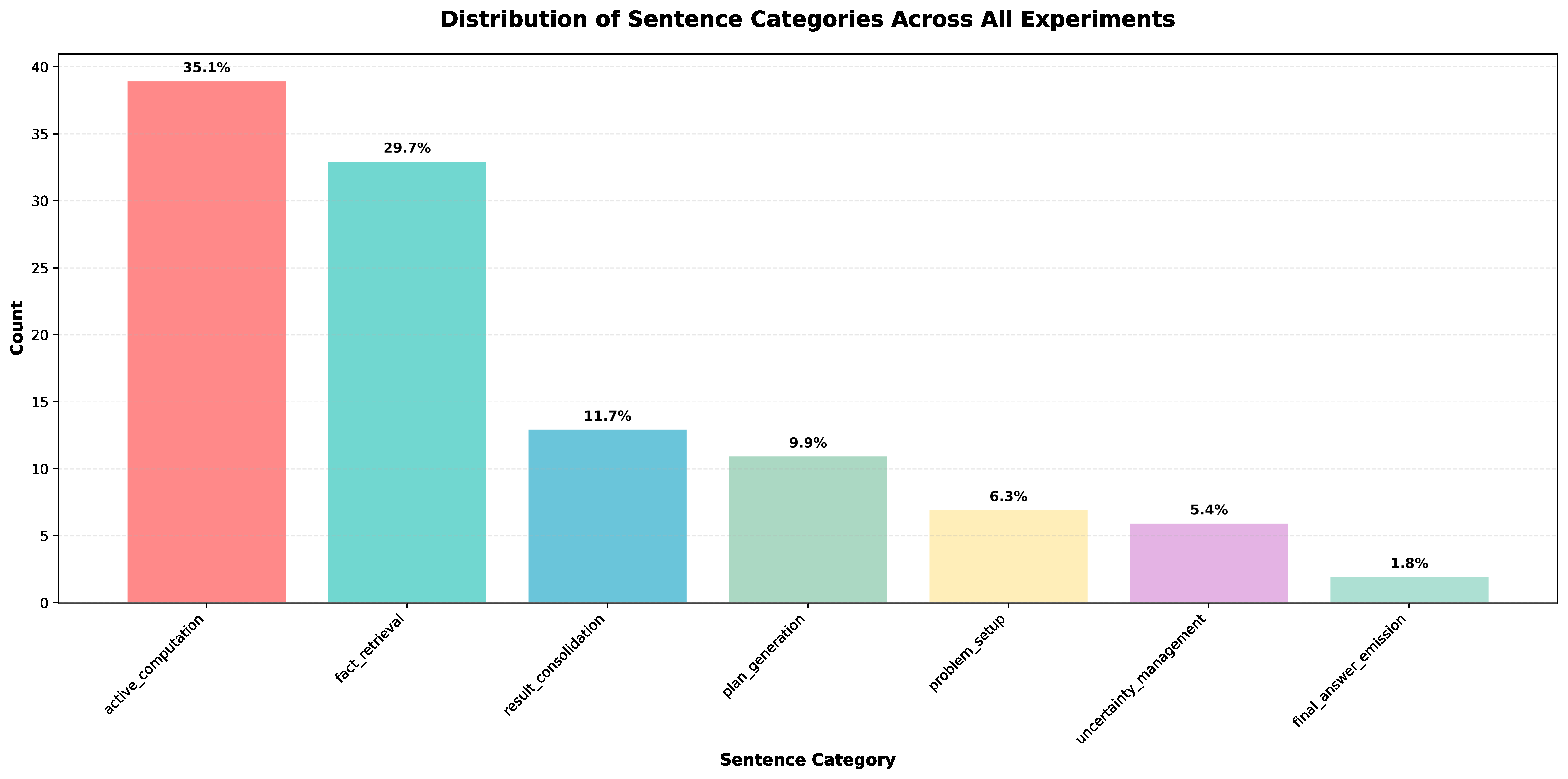}
    \caption{Sentence category vs counts across all experiments.}
    \label{fig:catdist}
\end{figure}

We report a similar sentence category distribution to \citet{bogdan2025thought} as shown in Figure \ref{fig:catdist}.

\section{Needle-In-A-Haystack: Additional Results}

\subsection{Information Flow Full Plots}

\begin{figure}[H]
    \centering
    \includegraphics[width=\linewidth]{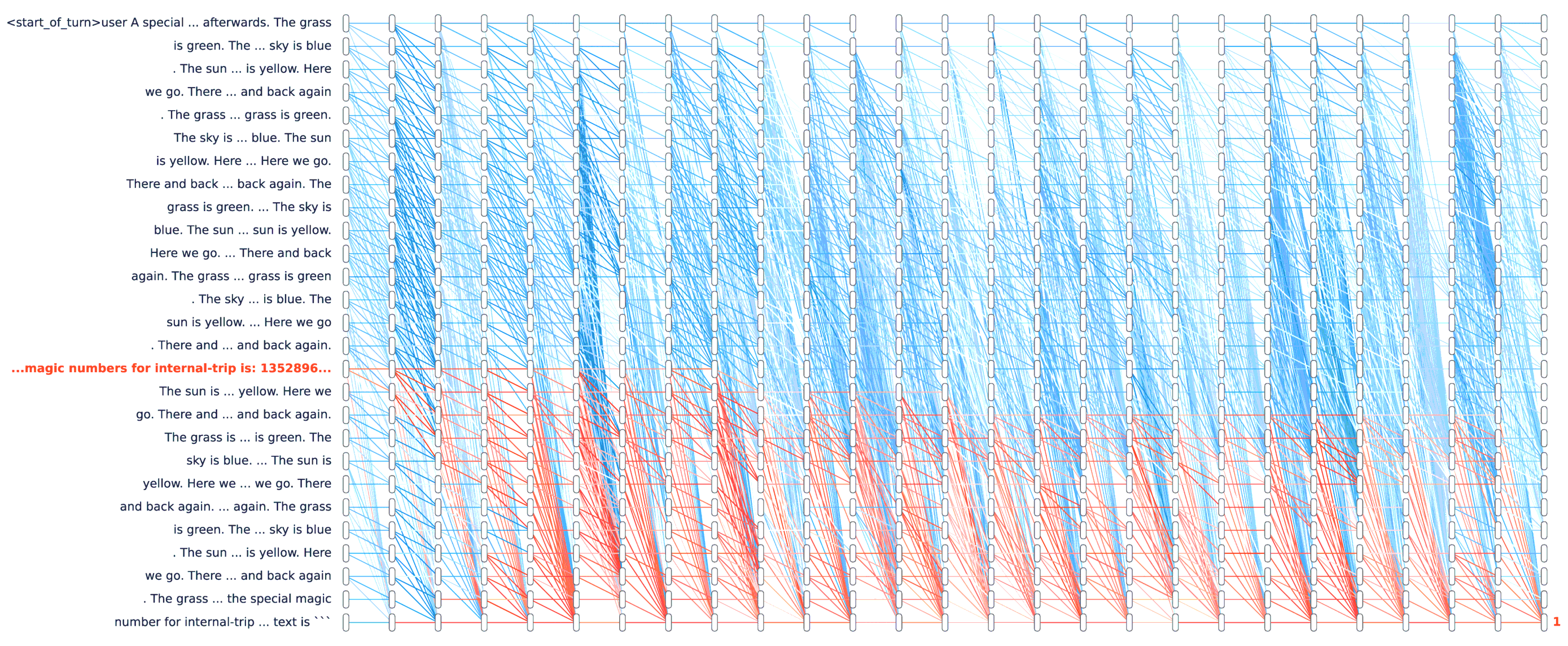}
    \caption{All information paths for a successful needle retrieval ($k=6$). We plot paths from the needle block to the output in red and the rest in blue.}
    \label{fig:graph}
\end{figure}

\begin{figure}[H]
    \centering
    \includegraphics[width=\linewidth]{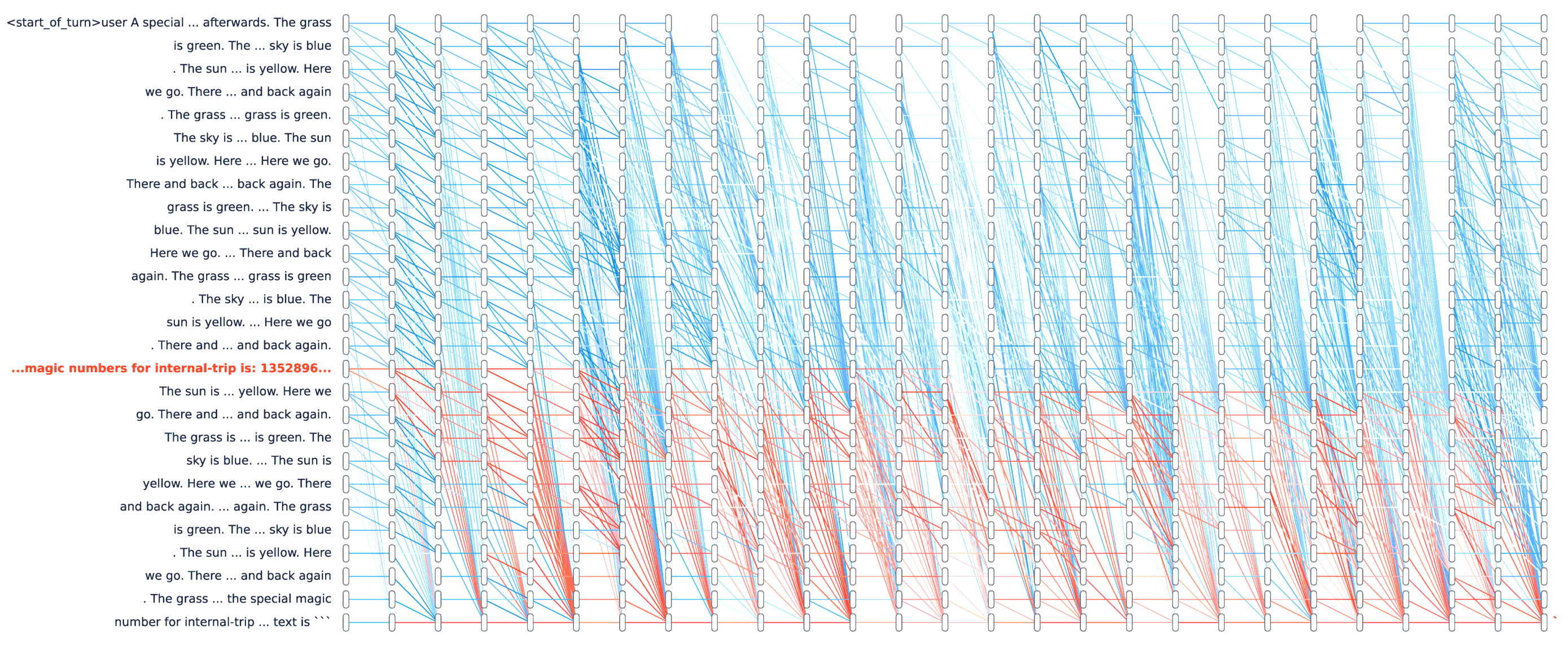}
    \caption{All information paths for a unsuccessful needle retrieval ($k=3$). We plot paths from the needle block to the output in red and the rest in blue.}
    \label{fig:graphfail}
\end{figure}

We report the information flow for a successful needle retrieval in Figure \ref{fig:graph} and unsuccessful retrieval in Figure \ref{fig:graphfail}.
\label{infapp}

\end{document}